\documentclass[journal]{IEEEtran}

\ifCLASSINFOpdf

\else
\fi
\usepackage{tabularx}
\usepackage{array}
\usepackage{graphicx}
\usepackage{subfigure}
\usepackage{float}
\usepackage{amsmath}
\usepackage{caption}
\usepackage{amssymb}
\usepackage{epstopdf}
\usepackage{dcolumn}
\usepackage{algorithmic}
\usepackage{algorithm}
\usepackage{pifont}
\usepackage{marvosym}
\usepackage{diagbox}
\usepackage{booktabs}
\usepackage{threeparttable}
\usepackage{multirow}
\usepackage{color}

\begin{document}

\title{Secure Artificial Intelligence of Things for Implicit Group Recommendations}

\author{Keping~Yu,~\IEEEmembership{Member,~IEEE,}
        Zhiwei~Guo,~\IEEEmembership{Member,~IEEE,}
        Yu~Shen,~
        Wei Wang,
        Jerry Chun-Wei Lin,~\IEEEmembership{Senior~Member,~IEEE, }
        Takuro Sato,~\IEEEmembership{Life Fellow,~IEEE}
\thanks{This work was supported in part by the State Language Commission Research Program of China under grant YB135-121, in part by the Science and Technology Research Program of Chongqing Municipal Education Commission under Grant KJQN202000805, in part by the Chongqing Natural Science Foundation of China under grant cstc2019jcyj-msxmX0747, in part by the Japan Society for the Promotion of Science (JSPS) Grants-in-Aid for Scientific Research (KAKENHI) under Grant JP18K18044, and in part by the Key Research Project of Chongqing Technology and Business University under grant ZDPTTD201917 and grant KFJJ2018071. (\emph{Corresponding author: Yu Shen})}
\thanks{Keping Yu is with Global Information and Telecommunication Institute, Waseda University, Tokyo 169-8555, Japan (e-mail: keping.yu@aoni.waseda.jp).}
\thanks{Zhiwei Guo is with School of Artificial Intelligence, Chongqing Technology and Business
University, Chongqing 400067, China (e-mail: zwguo@ctbu.edu.cn).}
\thanks{Yu Shen is with National Research Base of Intelligent Manufacturing Service, Chongqing Technology and Business
University, Chongqing 400067, China (e-mail: shenyu@ctbu.edu.cn).}
\thanks{Wei Wang is with School of Intelligent Systems Engineering, Sun Yat-Sen University, Shenzhen, China. (email: ehomewang@ieee.org)}
\thanks{Jerry Chun-Wei Lin is with the Department of Computer Science, Electrical Engineering and Mathematical Sciences, Western Norway University of Applied Sciences, 5063, Bergen, Norway (e-mail: jerrylin@ieee.org).}
\thanks{Takuro Sato is with Research Institute for Science and Engineering, Waseda University, Tokyo 169-8555, Japan (e-mail: t-sato@waseda.jp).}
}

\maketitle

\begin{abstract}
The emergence of Artificial Intelligence of Things (AIoT) has provided novel
insights for many social computing applications such as group recommender systems. As distance among people has been greatly shortened, it has been a more general demand to provide personalized services to groups instead of individuals. In order to capture group-level preference features from individuals, existing methods were mostly established via aggregation and face two aspects of challenges: secure data management workflow is absent, and implicit preference feedbacks is ignored. To tackle current difficulties, this paper proposes secure Artificial Intelligence of Things for implicit Group Recommendations (SAIoT-GR). As for hardware module,
a secure IoT structure is developed as the bottom support platform. As for software
module, collaborative Bayesian network model and non-cooperative game are can be
introduced as algorithms. Such a secure AIoT architecture is able to maximize the advantages of the two modules. In addition, a large number of experiments are carried out to evaluate the performance of the SAIoT-GR in terms of efficiency and robustness.
\end{abstract}


\IEEEpeerreviewmaketitle

\section{Introduction}

\IEEEPARstart{T}{he} constant prevalence of the Internet of Things (IoT) makes
it possible to construct a world in which all things are interconnected \cite{r1}.
At the same time, the emergence of artificial intelligence (AI) has also brought
novel vitality to multiple fields \cite{r2,r35}. The fusion of them yields an innovative
conception named Artificial Intelligence of Things (AIoT), a potentially promising
technology mode in the future \cite{r3}. Predictably, the AIoT may be adopted
to improve many ordinary industrial or commercial applications \cite{r8}, in which the
recommender systems (RSs) acts as a most typical one \cite{r4,r27}. In the context
of IoT, the continuous development of communication quality has substantially
boosted information transmission and exchange, yet also leading to remarkable
information overload issue \cite{r5}. Especially in the era of upcoming 5G, such
problem is likely to become more prominent in terms of IoT environment \cite{r7,r33}.
Nowadays, the RSs have been regarded as effective tools to deal with such issue,
and have gained considerable attention for some years \cite{r28}. The RSs
managed to provide personalized services to users, so that suitable
items that satisfy their preferences can be suggested to them \cite{r34}.
But existing RSs were mostly developed towards individual users \cite{r9}. But due to many conveniences brought by information techniques, more and more social activities tend to be arranged in the form of groups. As a result, recommending items to groups instead of individuals has also become a meaningful demand, yielding the application of group recommender systems (GRSs).
\begin{figure*}[tb]
\centering
  \includegraphics[width=12cm]{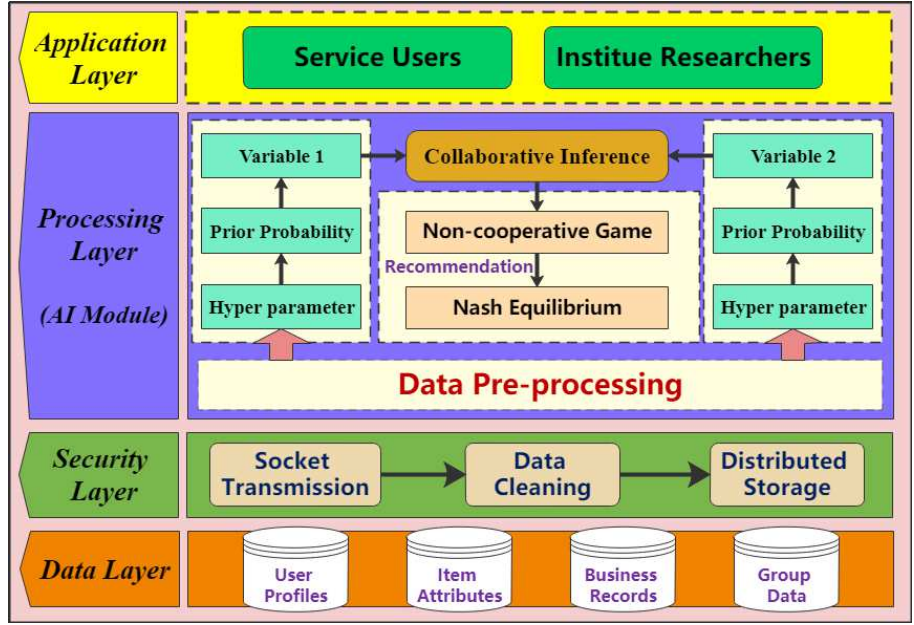}
\caption{Architecture for platform framework of the SAIoT-GR.}\label{fig-1 architecture}
\end{figure*}

Despite great progress has been obtained with respect to GRSs \cite{r10}-\cite{r26},
two aspects of challenges are still faced. For one thing, They were generally
constructed with the use of offline data, and lacked the ability to be adaptive
to changeable situations in real time. This is because the online sample training
highly depends on the reliable management of security guarantee and data integration.
In fact, real-time online management for data integration and security guarantee
is important under IoT environment that is filled with time-varying characteristics.
For another, existing GRSs were developed by investigating how to effectively aggregate
members into a group. However, almost all of these methods
just considered explicit preference feedbacks and failed to
handle the scenarios of implicit preference feedbacks. The
explicit feedbacks refer to those that are able to directly
reflect preference degree towards items, such as ratings.
In contrast, the implicit feedbacks refer to interactions between users and items
such as click records and purchase records, so that preference degree is indirectly
expressed. Undoubtedly, the explicit ones are much more intuitive and favourable for
knowledge discovery. But data with the form of implicit feedbacks is quite common
in real-world scenarios, as it usually remains difficult to acquire ideal data samples.

The aforementioned challenges can be tackled by designing a secure AIoT architecture
that contains two parts: hardware module and software module two parts. As for the hardware module, an IoT structure is developed as the bottom support platform. Its strong ability of data management and security guarantee will contribute a lot to the software module that is embedded into the hard module so that models for group recommendations can be re-optimized online. As for the software module, the collaborative Bayesian network (CBN) model can be introduced to simultaneously infer unknown preference features hidden inside implicit feedbacks. In other words, implicit feedbacks can be transformed into explicit ones through joint probabilistic inference. And then non-cooperative game theory can be utilized to search for optimal results for all the group members. Thus, this paper proposes \textbf{S}ecure
\textbf{A}rtificial \textbf{I}ntelligence \textbf{o}f \textbf{T}hings for \textbf{G}roup
\textbf{R}ecommendations, which is named SAIoT-GR for short. The fusion of secure
IoT bottom framework and AI algorithms is able to maximize their own advantages of
the two modules, and improves recommendation efficiency to the most extent. This work firstly considers formulating online data
management mechanism to establish an efficient GRS. In addition, it also firstly
investigates the group recommendation problem from the perspective of implicit
feedback. Major highlights of this paper are summed up as:
\begin{itemize}
\item It is recognized that existing researches concerning GRSs still face
two aspects of challenges. One is absence of real-time online management,
and the other is ignorance of situations of implicit feedbacks.
\item SAIoT-GR, composed of two main modules, is proposed to solve the above
two aspects of challenges, respectively. It well fuses a secure IoT framework
and AI algorithms.
\item Enough experiments on real-world datasets are
carried out to prove efficiency of the proposed SAIoT-GR.
\end{itemize}

Except introduction, this paper contains four main sections.
Section II, entitled system model, gives macroscopic design for both IoT architecture and the embedded
recommendation model. Among, the microscopic workflow of the recommendation
algorithm is described in Section III. The Section IV presents a series of
experiments which evaluate performance of the proposed SAIoT-GR. In Section
V, the paper is concluded.

\section{System Model}
This research manages to put forward SAIoT-GR to build up a novel GRS under
the situations of implicit feedbacks. This section firstly describes platform
framework of the designed SAIoT-GR, and then describes the specifically developed
CBN model that is embedded into the platform.
\subsection{Hardware Architecture}
Architecture of the designed SAIoT is illustrated in Fig.~\ref{fig-1 architecture},
which is actually the combination of a secure IoT platform and an embedded AI algorithm.
From the view of structural composition, it is composed of two main modules:
hardware module and software module. The former mainly refers to the IoT bottom
framework that offers data integration, and is endowed with a four-layer working
structure: data layer, security layer, processing layer and application layer. Among,
the processing layer directly implements group recommendations by carrying the
software module which is exactly the developed AI algorithm CBN model. All parts
of SAIoT-GR work jointly to constitute the newly proposed GRS. Three main
modules are described as the following contents:
\begin{figure*}[t]
\centering
  \includegraphics[width=16cm]{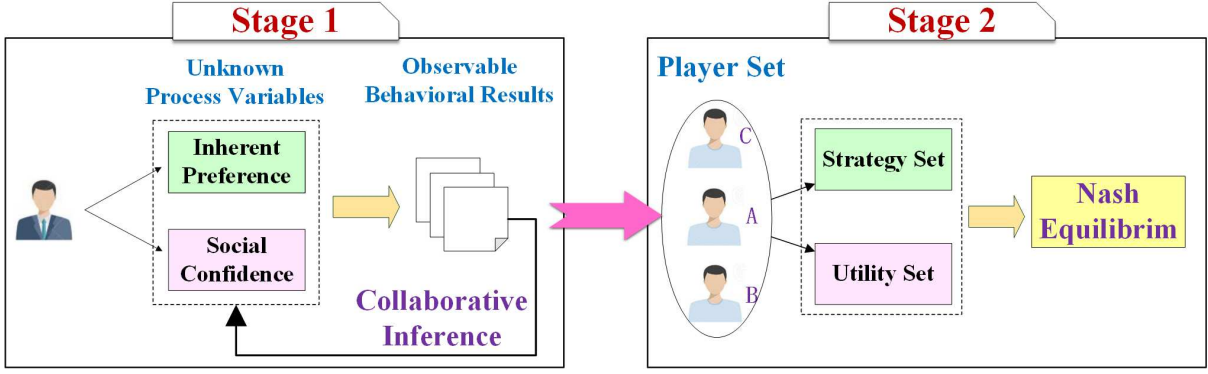}
\caption{Algorithmic workflow of the SAIoT-GR.}\label{fig-2 algorithm workflow}
\end{figure*}
\begin{itemize}
\item The data layer collects the original data and
connecting them to the SAIoT-GR. It contains several types of entities: user profiles,
item attributes, business records, and group information.
And all types of data related to group recommendations can be acquired and updated
in real-time under the environment of IoT.
\item The security layer is mainly responsible for some physical security management
of source data. Firstly, it provides links to transmit data from the data layer to
the following working layers. Secondly, it put forward unified authentication
mechanisms for multi-source heterogeneous data to ensure online information security.
Thirdly, it gives distributed storage space for source data.
\item The processing layer mainly implements the two-stage AI algorithm that is
embedded into SAIoT-GR. For one thing, it employs the CBN model to model the
generative process of interaction results between users and items, so that
collaborative inference of unknown preference feedbacks can be realized. For
another, a non-cooperative game is adopted to output recommendation results for
groups of users to satisfy their interests.
\end{itemize}

\subsection{Workflow of SAIoT-GR}
Enumerating $i$ from 1 to $\left| G \right|$ and $j$ from 1 to $Q$,
$u_i$ denotes user set of group $G$ and $v_j$ denotes the item set.
When user $u_i$ selects or consumes item $v_j$, there exists an
interaction record $B\left( {i,j} \right)$ between them. Hence,
$u_i$, $v_j$ and $B\left( {i,j} \right)$ are actually the initial
training data. To better classify contents of different items,
each item is assigned a topic indicator $g_j=d$, where $d$ ranges
from 1 to $D$. It should be noted that the topics here never have
specific meanings and are latent forms. The main task of this work
is to select appropriate items from another candidate item set,
so that demand of most users can be satisfied. The demand of user
$u_i$ towards item $v_j$ is determined by two aspects of factors:
inherent interest and social influence. The former is defined as
the initial preference feedback of users towards items, not being
influenced by social factors. The latter is defined as the influence
brought by social friends and independent from inherent interest.

Algorithmic workflow of the SAIoT-GR is demonstrated in Fig.~\ref{fig-2 algorithm workflow},
which contains two main components: collaborative inference module and recommendation
module. The former models forward decision processes of users as CBN model and
infers preference features via Gibbs sampling. The latter models the generation
of recommendation results as the non-cooperative game among group members, and
calculates results that maximizes utility of whole groups. Before them,  the
prevalent short text classification algorithm Twitter-LDA is utilized to
endow each item with a topic indicator to each item based upon associated textual contents.

\section{Methodology}

\subsection{Collaborative Bayesian Network}
$B\left( {i,j} \right)$, the interaction record between user $u_i$
and item $v_j$, is mainly determined by four aspects of factors:
\begin{itemize}
\item $g_j$, the topic indicator of the item;
\item $I\left( {i,d} \right)$, inherent interest of user $u_i$ about topic $d$;
\item $\pi\left( {i,d} \right)$, contribution rate of topic $d$ towards user $u_i$;
\item $S\left( {i,d} \right)$, social influence of user $u_i$ about item $d$.
\end{itemize}
It is assumed that inherent interest
$I\left( {k,i} \right)$ is drawn from the first Gaussian
distribution $I\left( {k,i} \right)\sim Gau\left( {{\mu _1},\sigma _1^2} \right)$,
where $\mu_1$ is mean and $\sigma_1^2$ is variance.
Similarly, it is also assumed that social confidence $S\left( {k,i} \right)$
is drawn from another Gaussian distribution
$S\left( {k,i} \right)\sim Gau\left( {{\mu _2},\sigma _2^2} \right)$, where $\mu_2$ is mean and $\sigma_2^2$ is variance.
Selection result of user $u_i$ on item $v_j$ is denoted as:
\begin{equation}
B\left( {i,j} \right) = \left\{ {\begin{array}{*{20}{c}}
{1,}&{\rm{selection}}\\
{0,}&{\rm{no~selection}}
\end{array}} \right.
\label{eq-1}
\end{equation}

For scenes where $B\left( {i,j} \right)=1$, their generative
processes can be deduced as conditional probability with
the use of logistic function:
\begin{equation}
\begin{split}
    & \mathcal{P}\left[ {B\left( {i,j} \right) = 1|{g_j=d},{\pi\left( {i,d} \right)},{I\left( {i,d} \right)},{S\left( {i,d} \right)}} \right]\\
={} & \frac{1}{{1 + \exp \left[ { - {\pi\left( {i,d} \right)}\cdot{I\left( {i,d} \right)} - {S\left( {i,d} \right)}} \right]}}
\end{split}
\label{eq-2}
\end{equation}
where $\pi\left( {i,d} \right)$ acts as the aforementioned
contribution rate between topic $d$ and user $u_i$. It is
calculated as:
\begin{equation}
\pi\left( {i,d} \right) = \frac{{m_i^{\left( d \right)}}}{{{m_i}}}
\label{eq-3}
\end{equation}
where $m_i$ is the amount of interactions of user $u_i$, and
${m_i^{\left( d \right)}}$ is the number of his interactions
related to topic $d$.
For scenes where $d_{j,i}=0$, generative process is expressed as:
\begin{equation}
\begin{split}
    & \mathcal{P}\left[ {B\left( {i,j} \right)=0|{g_j=d},{\pi\left( {i,d} \right)},{I\left( {i,d} \right)},{S\left( {i,d} \right)}} \right]\\
={} & 1 - \frac{1}{{1 + \exp \left[ { - {\pi\left( {i,d} \right)}\cdot{I\left( {i,d} \right)} - {S\left( {i,d} \right)}} \right]}}
\end{split}
\label{eq-4}
\end{equation}
Joint probability of all the generative processes is expressed as:
\begin{equation}
\begin{split}
    & \mathcal{P}\left[ {B\left( {i,j} \right)|{g_j=d},{\pi\left( {i,d} \right)},{I\left( {i,d} \right)},{S\left( {i,d} \right)}} \right]\\
={} & {\left\{ \frac{1}{{1 + \exp \left[ { - {\pi\left( {i,d} \right)}\cdot{I\left( {i,d} \right)} - {S\left( {i,d} \right)}} \right]}} \right\}^{\delta \left( {{d_{j,i}},1} \right)}}\\
    & \cdot {\left\{ {1 - \frac{1}{{1 + \exp \left[ { - {\pi\left( {i,d} \right)}\cdot{I\left( {i,d} \right)} - {S\left( {i,d} \right)}} \right]}}} \right\}^{\left[ {1 - \delta \left( {{d_{j,i}},1} \right)} \right]}}
\end{split}
\label{eq-5}
\end{equation}
where $\delta \left( {a,b} \right)$ is Kronecker delta represented as:
\begin{equation}
\delta \left( {a,b} \right) = \left\{ {\begin{array}{*{20}{c}}
{1,}&{a = b}\\
{0,}&{a \ne b}
\end{array}} \right.
\label{eq-6}
\end{equation}
Therefore, given the topic indicator $k$ and prior distributions,
the following formula can be deduced:
\begin{equation}
\begin{split}
    & \mathcal{P}\left[ {B\left( {i,j} \right),{\pi\left( {i,d} \right)},{I\left( {i,d} \right)},{S\left( {i,d} \right)}|{g_j} = d,{\mu _1},{\mu _2},{\sigma _1}^2,{\sigma _2}^2} \right]\\
={} & \mathcal{P}\left[ {{I\left( {i,d} \right)}|{\mu _1},{\sigma _1}^2} \right] \cdot \mathcal{P}\left[ {{S\left( {i,d} \right)}|{\mu _2},{\sigma _2}^2} \right]\\
    & \cdot \mathcal{P}\left[ {{B\left( {i,j} \right)}|{g_j=d},{\pi\left( {i,d} \right)},{I\left( {i,d} \right)},{S\left( {i,d} \right)}} \right]
\end{split}
\label{eq-7}
\end{equation}
Based on the above formula, learning goal for interactions can be summarized as
searching for minimization of the following objective function:
\begin{equation}
{J_{j,i}} =  - \log \mathcal{P}\left( {{d_{j,i}},{\pi _{k,i}},{I_{k,i}},{S_{k,i}}|{z_j} = k,{\mu _1},{\mu _2},{\sigma _1}^2,{\sigma _2}^2} \right)
\label{eq-8}
\end{equation}
Integrating the above two formulas leads to a new objective function:
\begin{equation}
  \begin{split}
    {J_{j,i}}  = & - \log \mathcal{P}\left[ {{I\left( {i,d} \right)}|{\mu _1},{\sigma _1}^2} \right] - \log \mathcal{P}\left[ {{S\left( {i,d} \right)}|{\mu _2},{\sigma _2}^2} \right]\\
    & - \log \mathcal{P}\left[ {B\left( {i,j} \right)|{g_j=d},{\pi\left( {i,d} \right)},{I\left( {i,d} \right)},{S\left( {i,d} \right)}} \right]
  \end{split}
\label{eq-9}
\end{equation}
And the formula can be rewritten as:
\begin{equation}
  \begin{split}
   {J_{j,i}}\propto & - \log \mathcal{P}\left[ {B\left( {i,j} \right)|{g_j=d},{\pi\left( {i,d} \right)},{I\left( {i,d} \right)},{S\left( {i,d} \right)}} \right]\\
    & - \frac{{{{\left[ {{I\left( {i,d} \right)} - {\mu _1}} \right]}^2}}}{{2{\sigma _1}^2}} - \frac{{{{\left[ {{S\left( {i,d} \right)} - {\mu _2}} \right]}^2}}}{{2{\sigma _2}^2}}
  \end{split}
\label{eq-10}
\end{equation}
The unknown variables $I\left( {i,d} \right)$ and $S\left( {i,d} \right)$
can be inferred through stochastic gradient descent method.

To solve the learning objective established above, the most
ordinary optimization approach named stochastic gradient descent
(SGD) is adopted here. Finally, the obtained process variables
are two vectors as the following forms:
\begin{equation}
I\left( {i,d} \right) = \left[ {{I\left( {i,1} \right)},{I\left( {i,2} \right)}, \cdots ,{I\left( {i,D} \right)}} \right]\left| {_{i \in \left[ {1,\left| G \right|} \right]}} \right.
\label{eq-12}
\end{equation}
\begin{equation}
S\left( {i,d} \right) = \left[ {{S\left( {i,1} \right)},{S\left( {i,2} \right)}, \cdots ,{S\left( {i,D} \right)}} \right]\left| {_{i \in \left[ {1,\left| G \right|} \right]}} \right.
\label{eq-13}
\end{equation}

\subsection{Recommendation}
When making group recommendations, intuitively, it is expected to
analyze the cooperation relations and competition intention of
group members. And naturally, recommendation results to a group
need to be item sets rather than single items. Thus, the idea of non-cooperative game is introduced here to allocate recommended
item sets to the whole group. When the equilibrium state is reached,
corresponding item sets can be viewed as the optimal allocation
for the whole group.

As for the construction of game process, three elements are
included: player set, strategy set and utility set. Generalized
to a group $G$, all of its members are regarded to take part
in the game process. Therefore, all of its members are actually
players of the game. In other words, the player set is the set
of group members whose size is $\left| G \right|$. During the
game process, each player will be asked to select one topic
indicator as his strategy $t_i$. The strategies of all the
group members constitute the strategy set. The size of strategy
set is $\left| G \right|$. It is assumed that there exists a
utility value $H_i$ for user $u_i$ to benefit from his selected
strategy. Similarly, the size of utility set also equals to
$\left| G \right|$.

\begin{figure}[tb]
\centering
  \includegraphics[width=7.5cm]{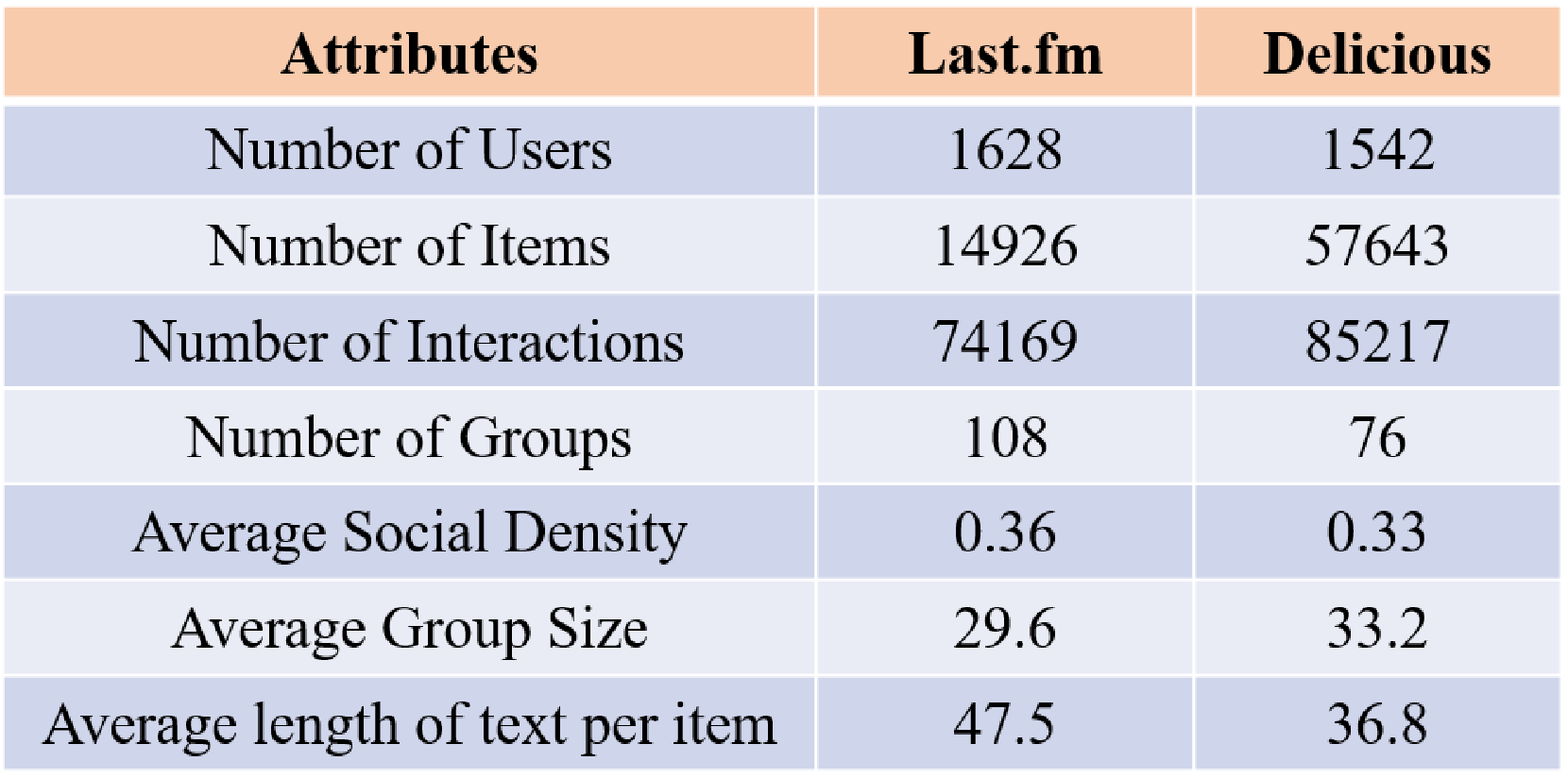}
\caption{Architecture for platform framework of the SAIoT-GR.}\label{fig-dataset statistics}
\end{figure}
In order to unify the range of all the utility values in the
utility set, it is supposed to normalized all the $I\left( {i,d} \right)$
and $S\left( {i,d} \right)$ into values with range $\left[ {0,1} \right]$.
Let $I_N\left( {i,d} \right)$ and $S_N\left( {i,d} \right)$
denote transformed forms of $I\left( {i,d} \right)$ and
$S\left( {i,d} \right)$. Given this, the final profit of
user $u_i$ can be computed as:
\begin{equation}
{X_i} = \frac{EXP\left[{s_i} \cdot I_N\left( {i,d} \right)\right]}{Count\left( {{s_i}} \right)}
\label{eq-16}
\end{equation}
where $EXP\left( \cdot \right)$ denotes the expectation operator,
and $Count\left( {{s_i}} \right)$ counts the number of players
whose strategies are $s_i$. At the same time, the cost of user $u_i$
born from $s_i$ is represented as:
\begin{equation}
{M_i} = {\eta _1} \cdot {\left[ {S_N\left( {i,d} \right)  + 1 - I_N\left( {i,d} \right) } \right]^n }
\label{eq-17}
\end{equation}
where $n$ is a model parameter. Offset by cost, utility finally
acquired by $u_i$ is computed as:
\begin{equation}
{H_i} = {\eta _2} \cdot \left( {X_i - M_i} \right)
\label{eq-18}
\end{equation}
where $\eta_1$ and $\eta_2$ are trade-off parameters.

Let ${\Omega ^*} = \{ s_1^*,s_2^*, \cdots ,s_{\left| G \right|}^*\} $
denote the strategy set under the status where Nash equilibrium
is reached. From the perspective of game theory, all of the players
cannot obtain larger utility by changing their strategies. At such
state, the following condition can be satisfied:
\begin{equation}
{H_i}(s_i^*,s_{ - i}^*) \le {H_i}({s_i},s_{ - i}^*)
\label{eq-19}
\end{equation}
where $s_{ - i}$ denotes strategy set from players except player $u_i$, and
$s_{ - i}^*$ denotes denotes such strategy set under status of Nash equilibrium.
When all the elements of ${\Omega ^*}$ are inferred, the contents
that are recommended to group $G$ are denoted as:
\begin{equation}
{R_G} = \left\{ {{A_1},{A_2}, \cdots ,{A_K}} \right\}
\label{eq-20}
\end{equation}
where $A_k$ is the ratio of items with topic indicator $k$.
In stead of specific items, recommendation results here take
the form of ratios of each topic indicator.

\begin{figure*}[t]
\centering
  \subfigure[SAIoT-GR]
  {
    \includegraphics[width=7cm]{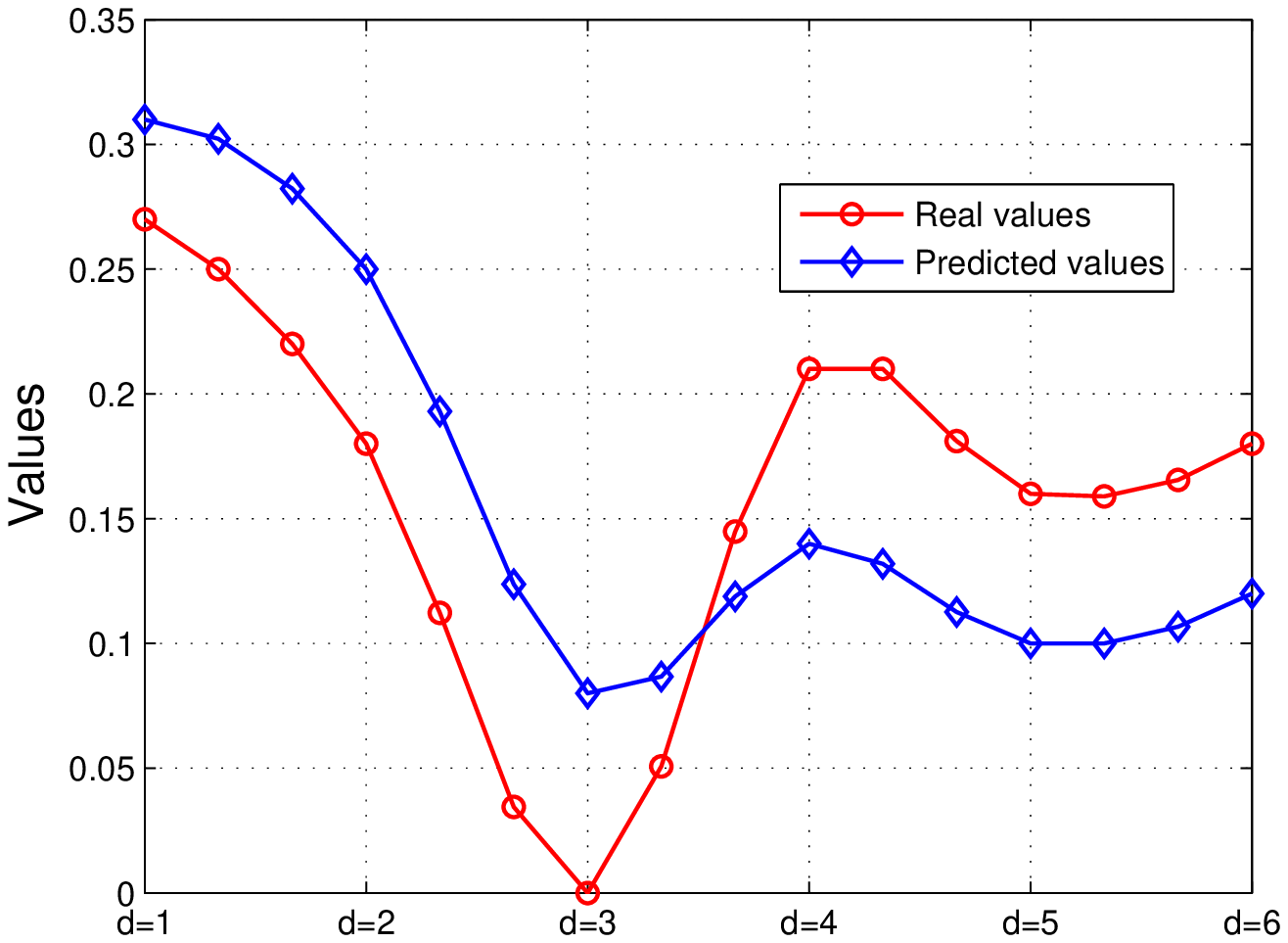}
    \label{fig-4a}
  }
  \subfigure[Frequency]
  {
    \includegraphics[width=7cm]{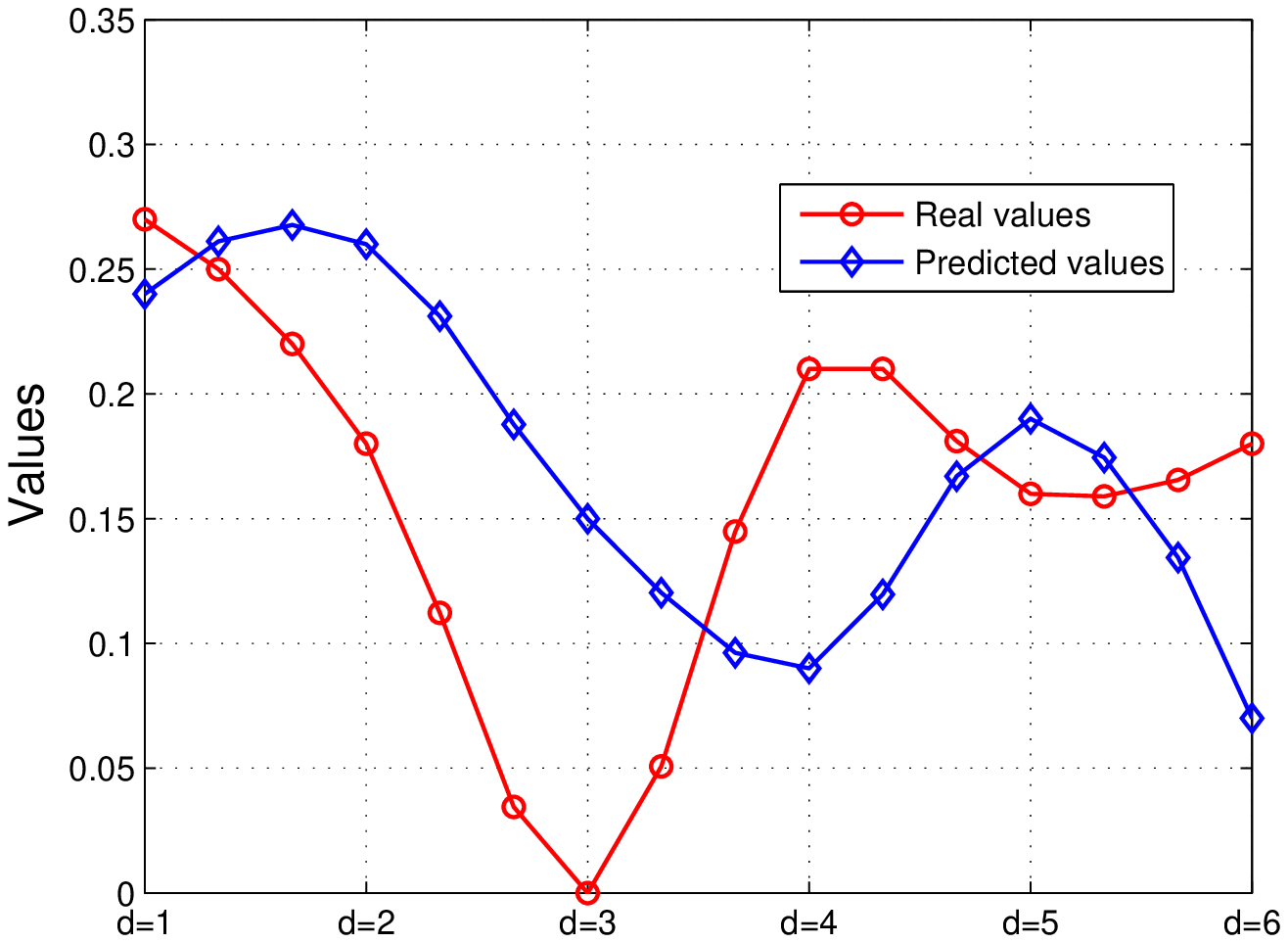}
    \label{fig-4b}
  }
  \subfigure[RanGroup]
  {
    \includegraphics[width=7cm]{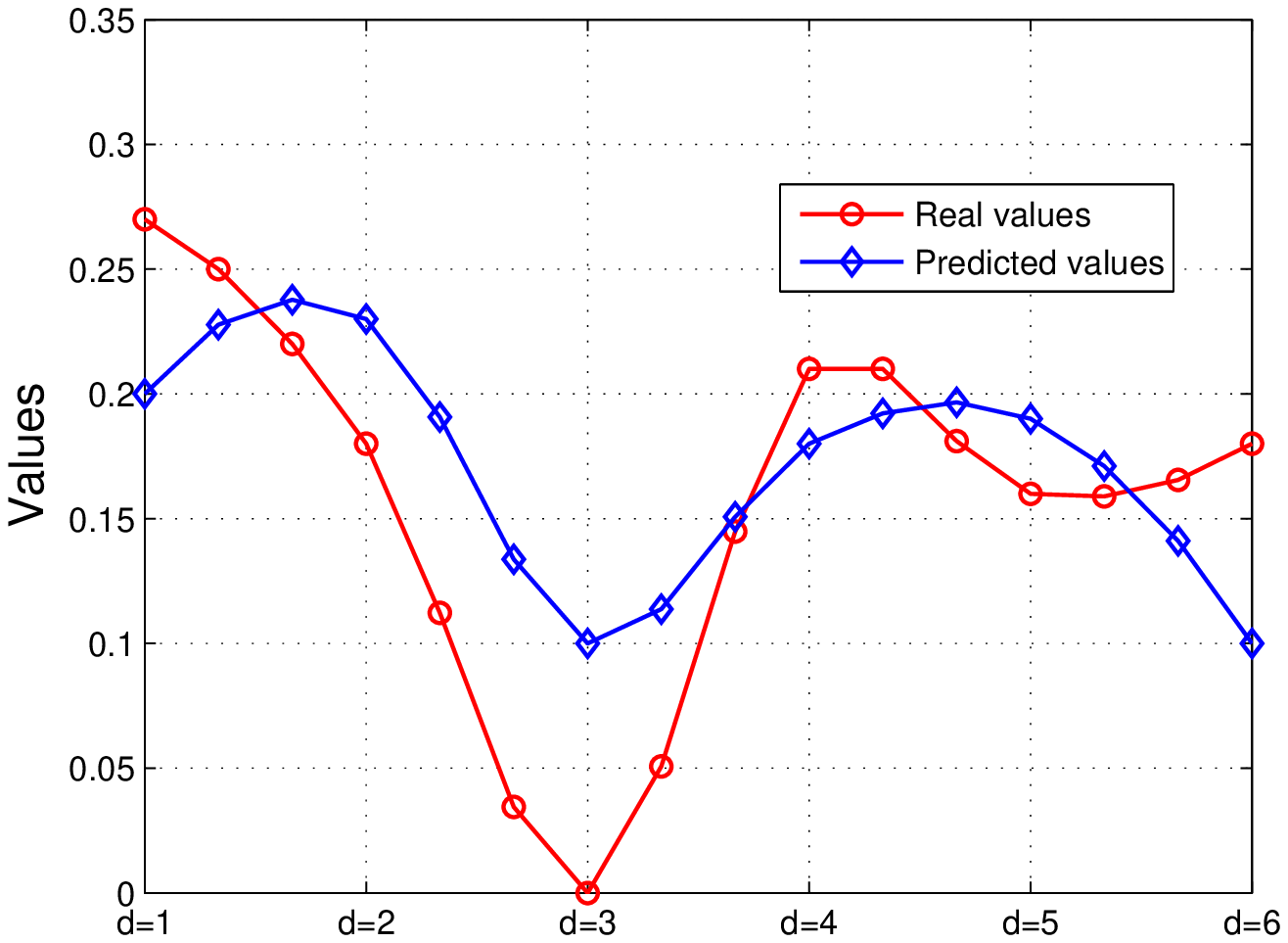}
    \label{fig-4c}
  }
  \subfigure[ConfiMF]
  {
    \includegraphics[width=7cm]{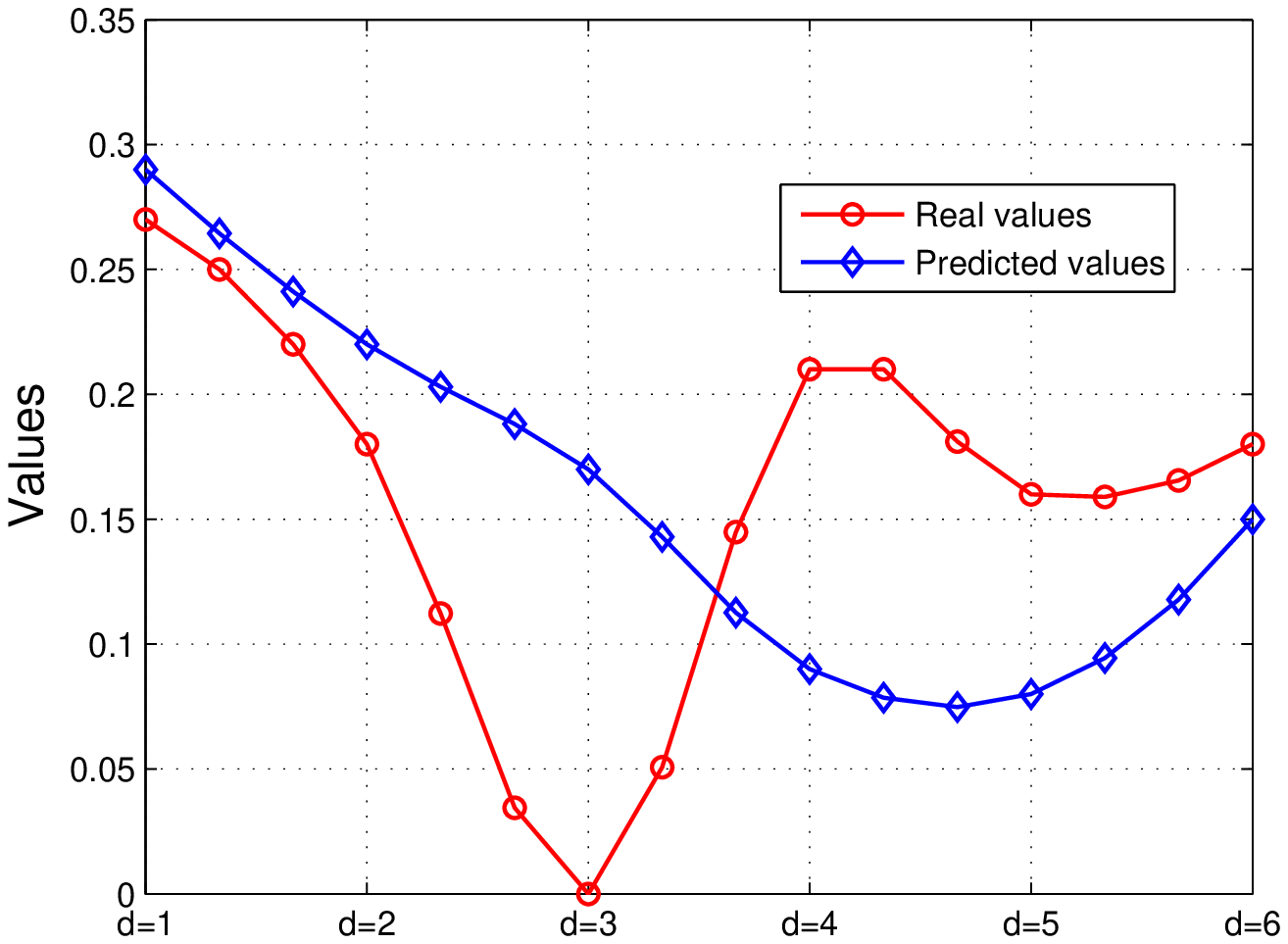}
    \label{fig-4d}
  }
  \subfigure[ContextMF]
  {
    \includegraphics[width=7cm]{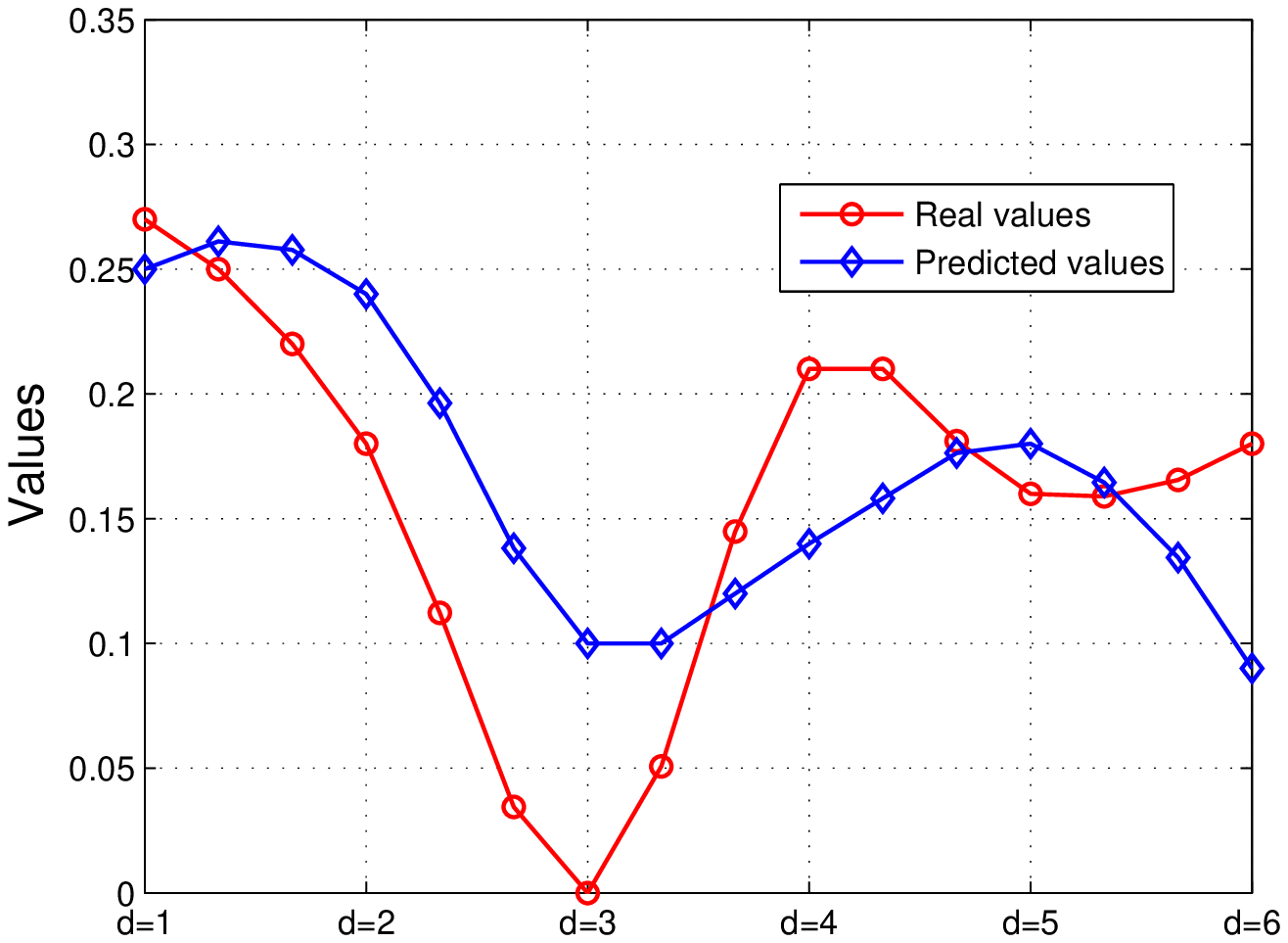}
    \label{fig-4e}
  }
  \subfigure[FreGroup]
  {
    \includegraphics[width=7cm]{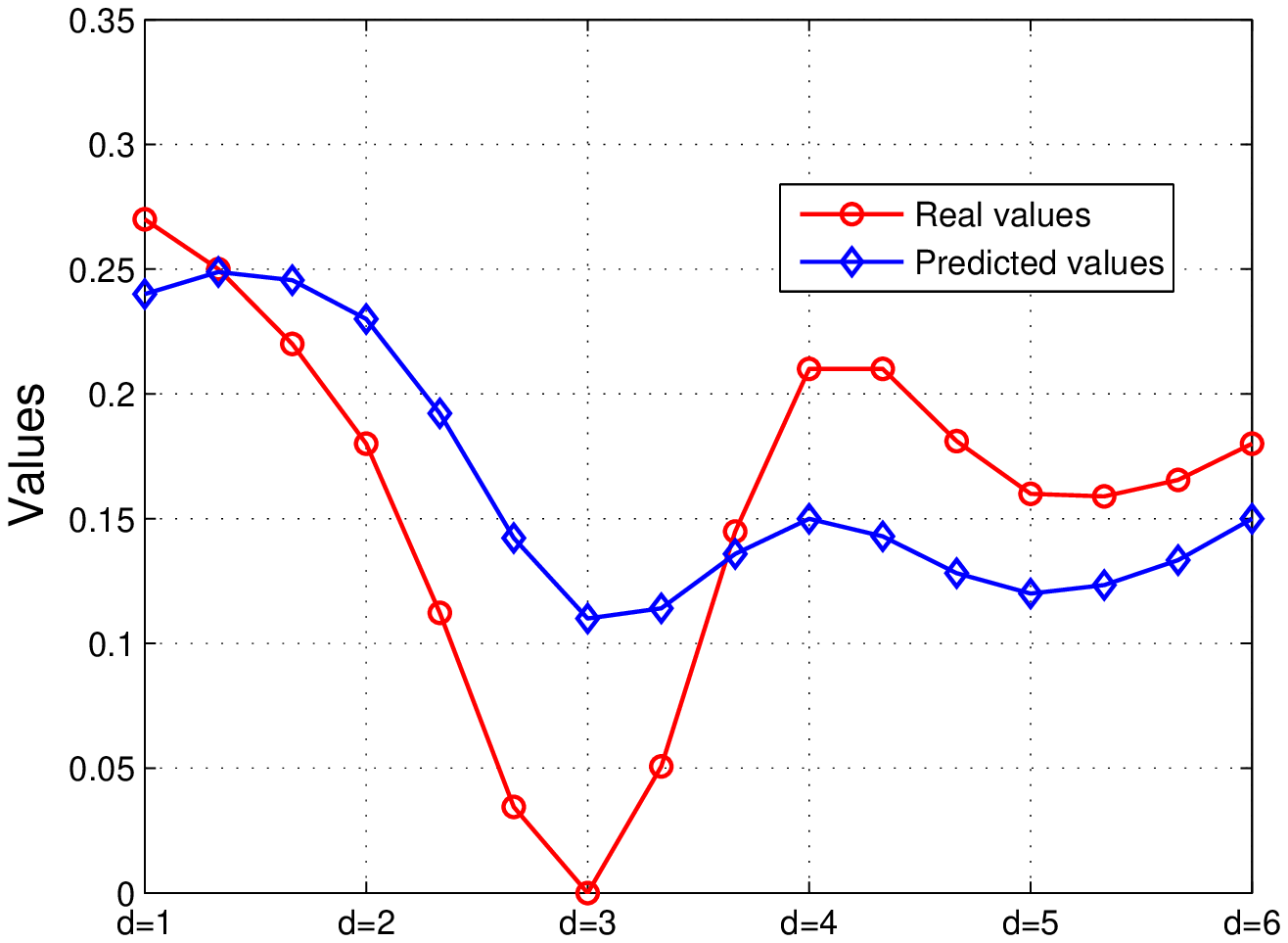}
    \label{fig-4f}
  }
\caption{Comparison between predicted topic distributions and real ones on the Last.fm.}
\label{fig-4 distribution comparision-last.fm}
\end{figure*}

\section{Experiments and Analysis}

\begin{figure*}[t]
\centering
  \subfigure[SAIoT-GR]
  {
    \includegraphics[width=7cm]{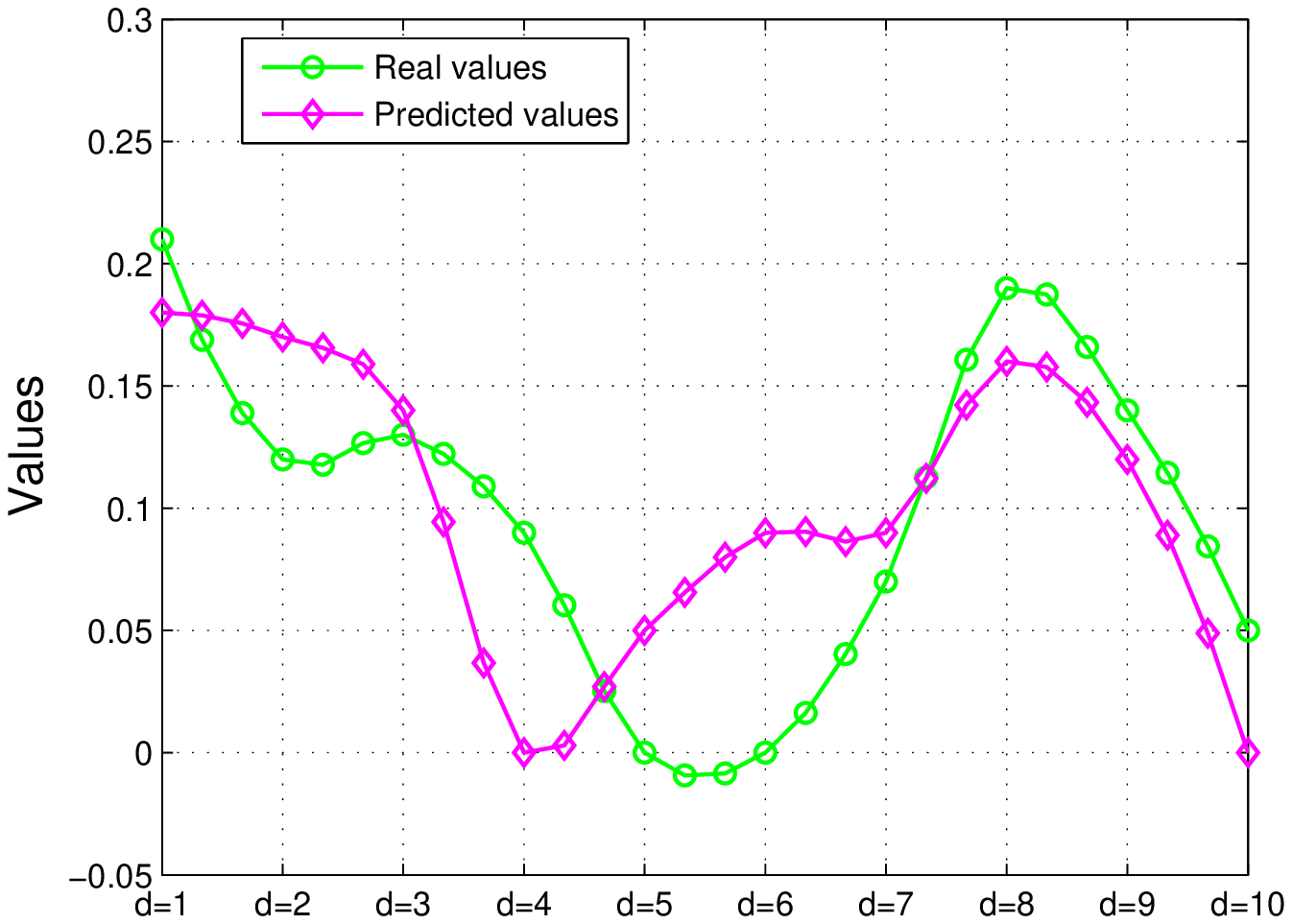}
    \label{fig-5a}
  }
  \subfigure[Frequency]
  {
    \includegraphics[width=7cm]{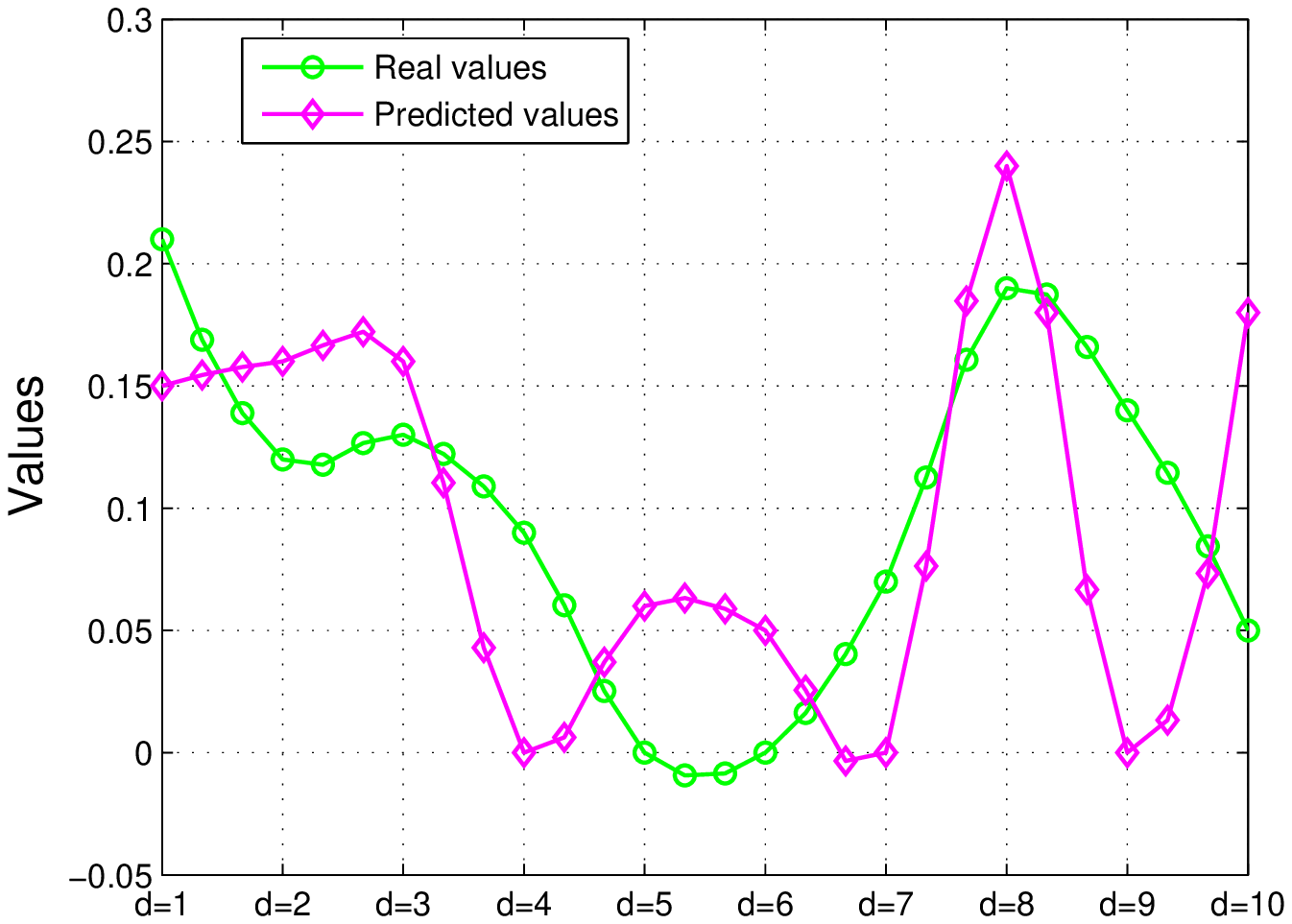}
    \label{fig-5b}
  }
  \subfigure[RanGroup]
  {
    \includegraphics[width=7cm]{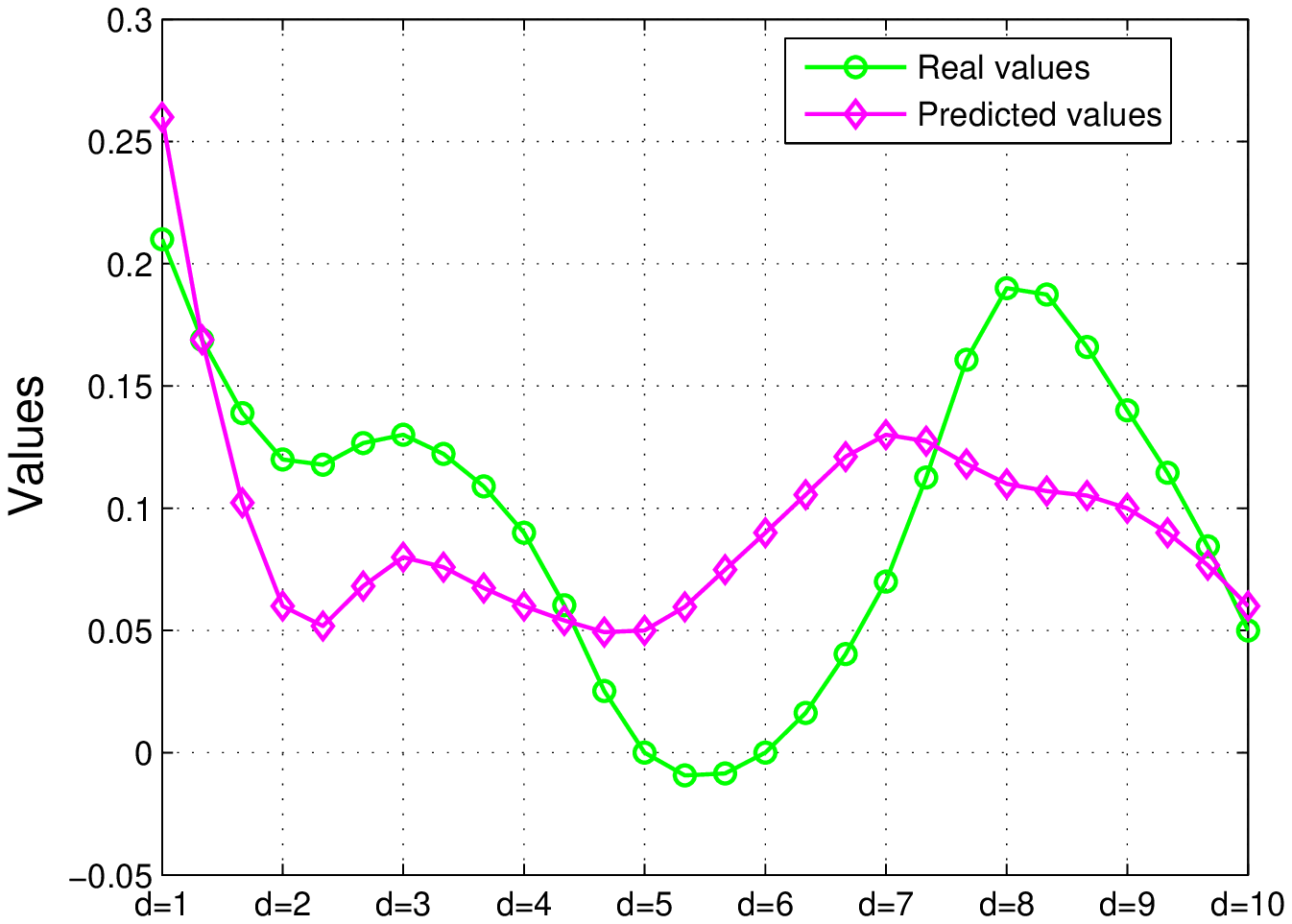}
    \label{fig-5c}
  }
  \subfigure[ConfiMF]
  {
    \includegraphics[width=7cm]{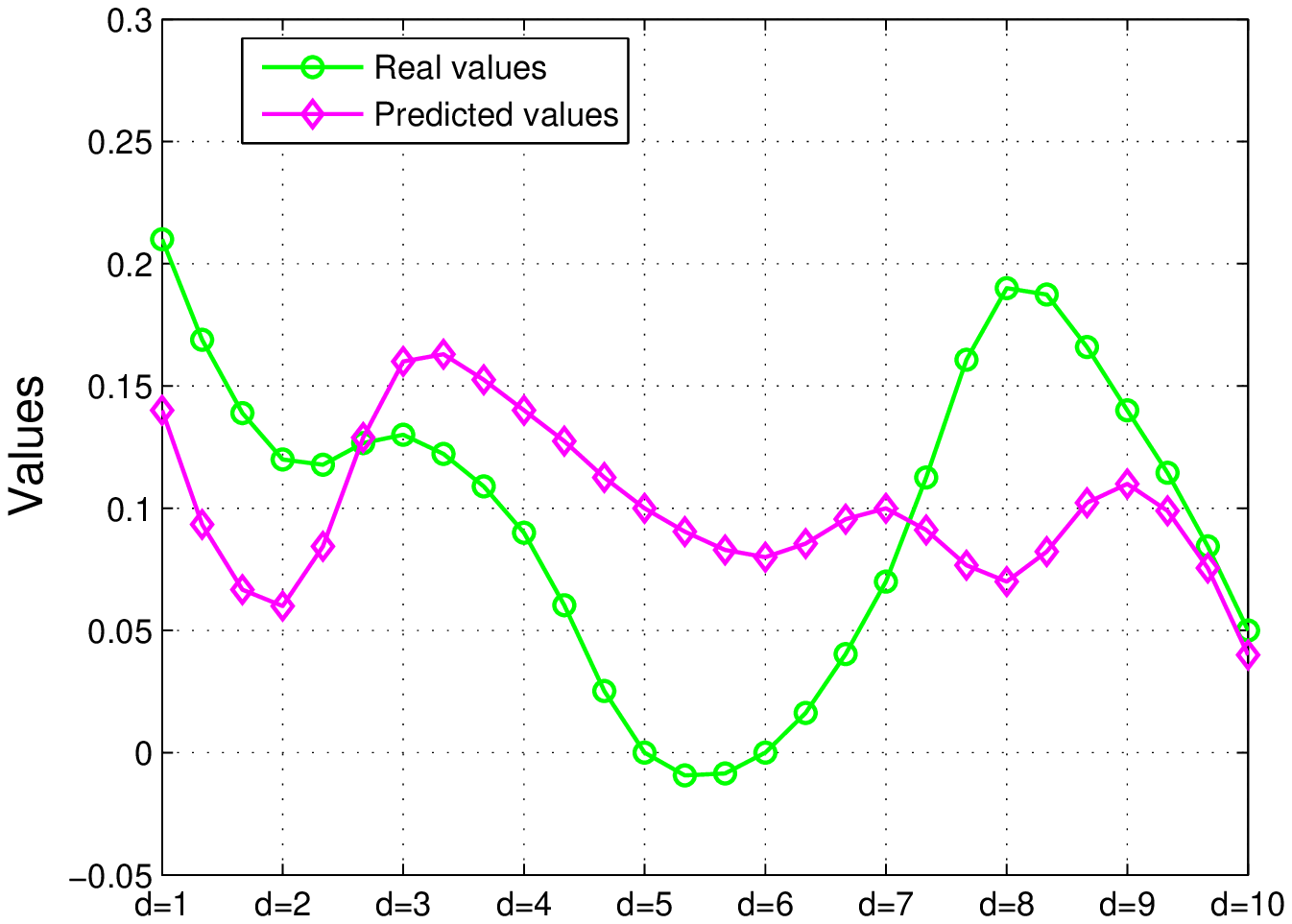}
    \label{fig-5d}
  }
  \subfigure[ContextMF]
  {
    \includegraphics[width=7cm]{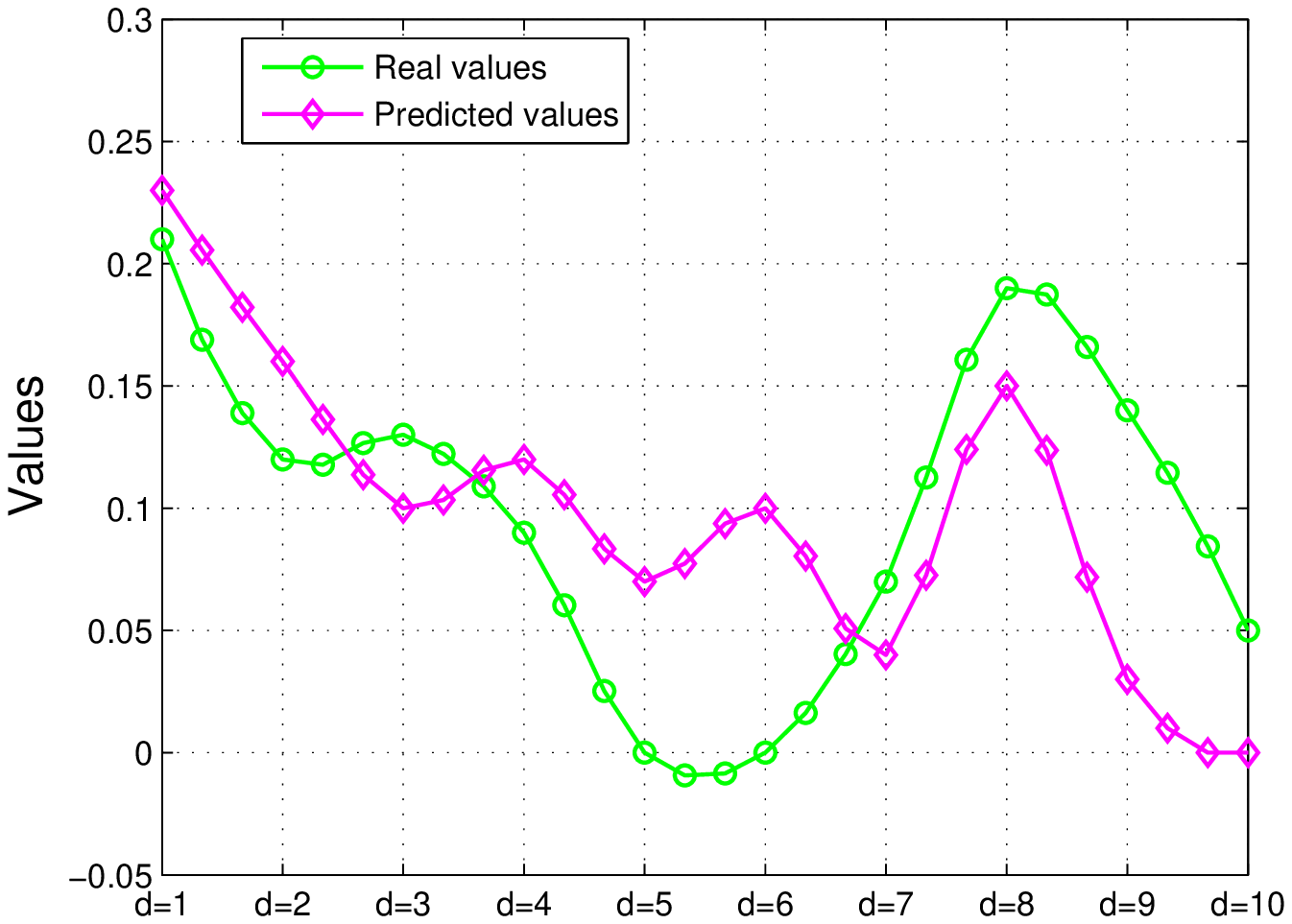}
    \label{fig-5e}
  }
  \subfigure[FreGroup]
  {
    \includegraphics[width=7cm]{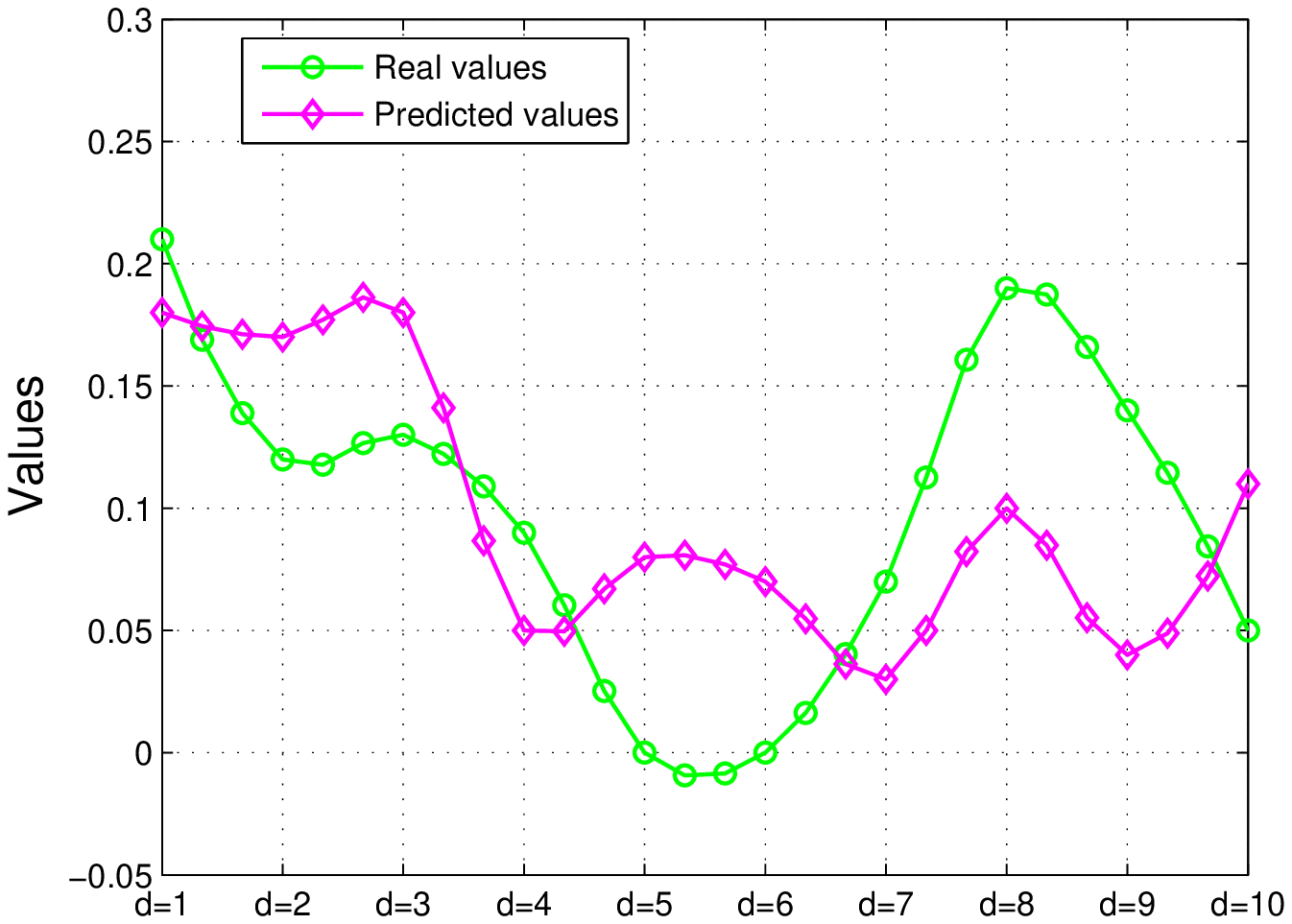}
    \label{fig-5f}
  }
\caption{Comparison between predicted topic distributions and real ones on the Delicious.}
\label{fig-5 distribution comparision-delicious}
\end{figure*}
\subsection{Datasets}
The performance of recommender systems is generally evaluated
on benchmark datasets. However, specific datasets for situations
of group recommendations are not publicly available since now.
Relevant researches universally artificially constructed experimental
datasets from those for situations of individual recommendations.
Here, two prevalent datasets that are commonly used for individual
recommendations, are selected for this purpose. As for group construction,
it is expected to randomly select some users to constitute some groups according to group size. Their statistical values can be found in
Fig.~\ref{fig-dataset statistics}, and they are briefly described as:

1) \textbf{Last.fm dataset: }Last.fm \footnote{https://www.last.fm/} is an
online music community with online social function, in which users can join
social communities and tag things they are interested in. The dataset was
collected by the team of Millennium song dataset through APIs provided by the
website.

2) \textbf{Delicious dataset: }Delicious \footnote{https://www.delicious.com}
is the largest bookmark sharing community of the Internet and also an online
social platform. Similar to Last.fm, users are allowed to release tags and
participate social communities. And the dataset is also a part of the Millennium
song dataset.

To reduce the effect brought by data sparsity,
inactive users with their interaction records inside the two datasets
are required to be filtered out, especially the users whose
number of interactions is small. Besides, some social density
is also required to ensure the richness of data. It is assumed
that the minimum social density inside a group must not be below
0.25. The social density is calculated as
${{\left( {2{E_G}} \right)} \mathord{\left/
 {\vphantom {{\left( {2{E_G}} \right)} {\left( {\left| G \right| \cdot \left| {G - 1} \right|} \right)}}} \right.
 \kern-\nulldelimiterspace} {\left( {\left| G \right| \cdot \left| {G - 1} \right|} \right)}}$, where $E_G$ is the number of social
relations that can be observed in group $G$.

\begin{table*}[!ht]
  \linespread{1.7}
  \centering
  \fontsize{10}{9}\selectfont
  \begin{threeparttable}
  \caption{Comparisons between SAIoT-GR and baselines with respect to six distances (Last.fm dataset).}
  \setlength{\tabcolsep}{3.5mm}
  \label{tab-2 six distances-last.fm}
    \begin{tabular}{ccccccc}
    \toprule
    \toprule
    \multirow{2}{*}{Method}&
    \multicolumn{6}{c}{Evaluation Metrics}\cr
    \cmidrule(lr){2-7}
                &EucDist   &ManDist  &CheDist  &CorDist  &MAEDist  &MSEDist   \cr
    \midrule
    SAIoT-GR    & \bf{0.1581}   & \bf{0.3800}  & \bf{0.0800}  & \bf{0.3008}  & 0.0633  & \bf{0.0042} \cr
    Frequency   & 0.2392   & 0.5200  & 0.1500  & 0.8050  & 0.0867  & 0.0095 \cr
    RanGroup    & 0.1600   & 0.3600  & 0.1000  & 0.3904  & 0.0653  & 0.0043 \cr
    ConfiMF     & 0.2293   & 0.4600  & 0.1700  & 0.7237  & 0.0767  & 0.0088 \cr
    ContextMF   & 0.1655   & 0.3600  & 0.1000  & 0.4046  & \bf{0.0600}  & 0.0046 \cr
    FreGroup    & 0.1969   & 0.4600  & 0.1600  & 0.5462  & 0.0676  & 0.0062 \cr
    \bottomrule
    \bottomrule
    \end{tabular}
    \end{threeparttable}
\end{table*}

\begin{table*}[!ht]
  \linespread{1.7}
  \centering
  \fontsize{10}{9}\selectfont
  \begin{threeparttable}
  \caption{Comparisons between SAIoT-GR and baselines with respect to six distances (Delicious dataset).}
  \setlength{\tabcolsep}{3.5mm}
  \label{tab-3 six distances-delicious}
    \begin{tabular}{ccccccc}
    \toprule
    \toprule
    \multirow{2}{*}{Method}&
    \multicolumn{6}{c}{Evaluation Metrics}\cr
    \cmidrule(lr){2-7}
                &EucDist   &ManDist  &CheDist  &CorDist  &MAEDist  &MSEDist   \cr
    \midrule
    SAIoT-GR    & \bf{0.1625}   & \bf{0.4400}  & \bf{0.0900}  & \bf{0.3040}  & \bf{0.0440}  & \bf{0.0026} \cr
    Frequency   & 0.2534   & 0.7200  & 0.1400  & 0.5424  & 0.0720  & 0.0064 \cr
    RanGroup    & 0.1783   & 0.5200  & 0.0900  & 0.3856  & 0.0520  & 0.0032 \cr
    ConfiMF     & 0.2112   & 0.5800  & 0.1200  & 0.6919  & 0.0580  & 0.0045 \cr
    ContextMF   & 0.1892   & 0.5200  & 0.1100  & 0.3999  & 0.0520  & 0.0036 \cr
    FreGroup    & 0.1858   & 0.6200  & 0.1500  & 0.4547  & 0.0560  & 0.0042 \cr
    \bottomrule
    \bottomrule
    \end{tabular}
    \end{threeparttable}
\end{table*}
\subsection{Experimental Settings}

All the experiments can be divided into two parts: efficiency assessment and
stability assessment. The former part measures precision of recommendation
results by comparing predicted results with real ones. For the latter, a
combination of parameters are changed into multiple groups to measure
robustness of the SAIoT-GR. Specifically, recommendation precision is
determined by the distance between real proportion distributions of topics
and predicted ones. To measure such distance, six different
distances are selected as metrics: Euclidean distance (EucDist),
Manhattan distance (ManDist), Chebyshev distance (CheDist), Correlation
distance (CorDist), mean absolute error (MAEDist), and Mean-square error
(MSEDist). The detailed definitions of the six distances can be found in one
of our previously published work \cite{r32}.

And five implicit feedbacks-based recommendation methods are selected as baselines,
and are briefly introduced as follows:

1) \textbf{Frequency: }It measures the preference features of group members by counting the frequency of different topics. And preferences of members are aggregated into group preferences.

2) \textbf{ConfiMF: }The preference confidence is regarded as explicit preference
feedback score which can be calculated through interaction frequency. Then, group
preference feedback is obtained through preference aggregation \cite{r29}.

3) \textbf{ContextMF: }It is assumed that decision process of a user towards an
item depends on two contextual information: user context and item context \cite{r30}.

4) \textbf{RanGroup: }It is a graph theory-based group recommender system and
produces recommendation results through idea of random work \cite{r31}.

5) \textbf{FreGroup: }It is a self-defined recommendation method that aggregates
interactions of members into interactions of groups. The, the Frequency method
is utilized to generate recommendation results for groups of users.

And some key parameters are required to
be specially set with respect to two datasets. For the Last.fm
dataset, mean values of two Gaussian distributions are set to
45 and 12 separately, variance values of two Gaussian distributions
are set to 70 and 30 separately. For the Delicious dataset, two
mean values are set to 45 and 10 separately, and two variance
values are set to 75 and 25 separately. Learning rate and the
convergence threshold of SAIoT-GR in experiments are set to
0.01 and 0.001, respectively. In the game process used for
generating, two trade-off parameters $\eta_1$ and $\eta_2$ in Eq.~(\ref{eq-17}) and (\ref{eq-18}), are set to 0.6 and 0.4.
And for the proportion of training data, it is set to 70\% by
default and can be changed during experimental processes.
The total number of topic indicators in Last.fm and Delicious
is set to 6 and 10, respectively.

\begin{figure*}[!ht]
\centering
  \subfigure[$\mu_1$ and $\sigma_1^2$]
  {
    \includegraphics[width=6cm]{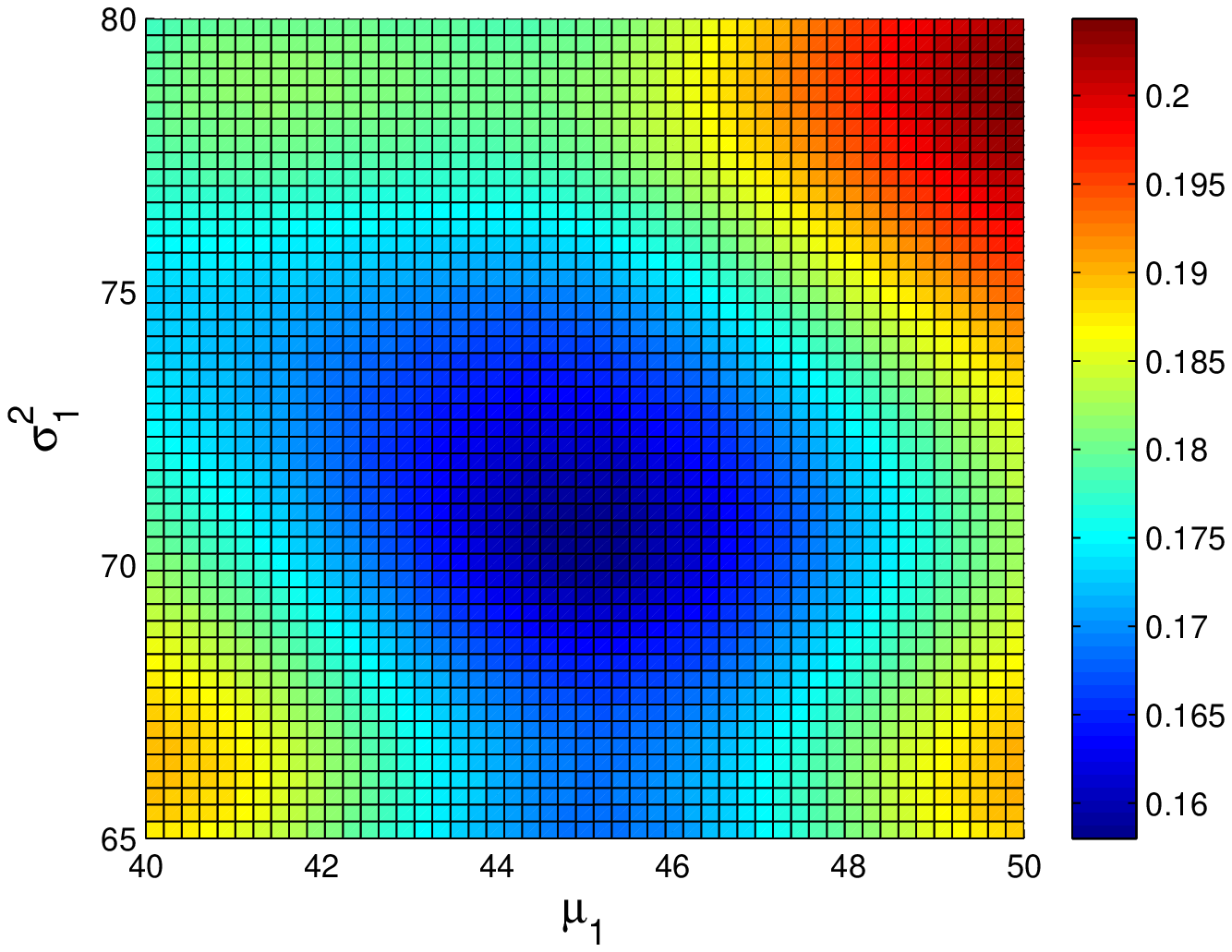}
    \label{fig-7a}
  }
  \subfigure[$\mu_2$ and $\sigma_2^2$]
  {
    \includegraphics[width=6cm]{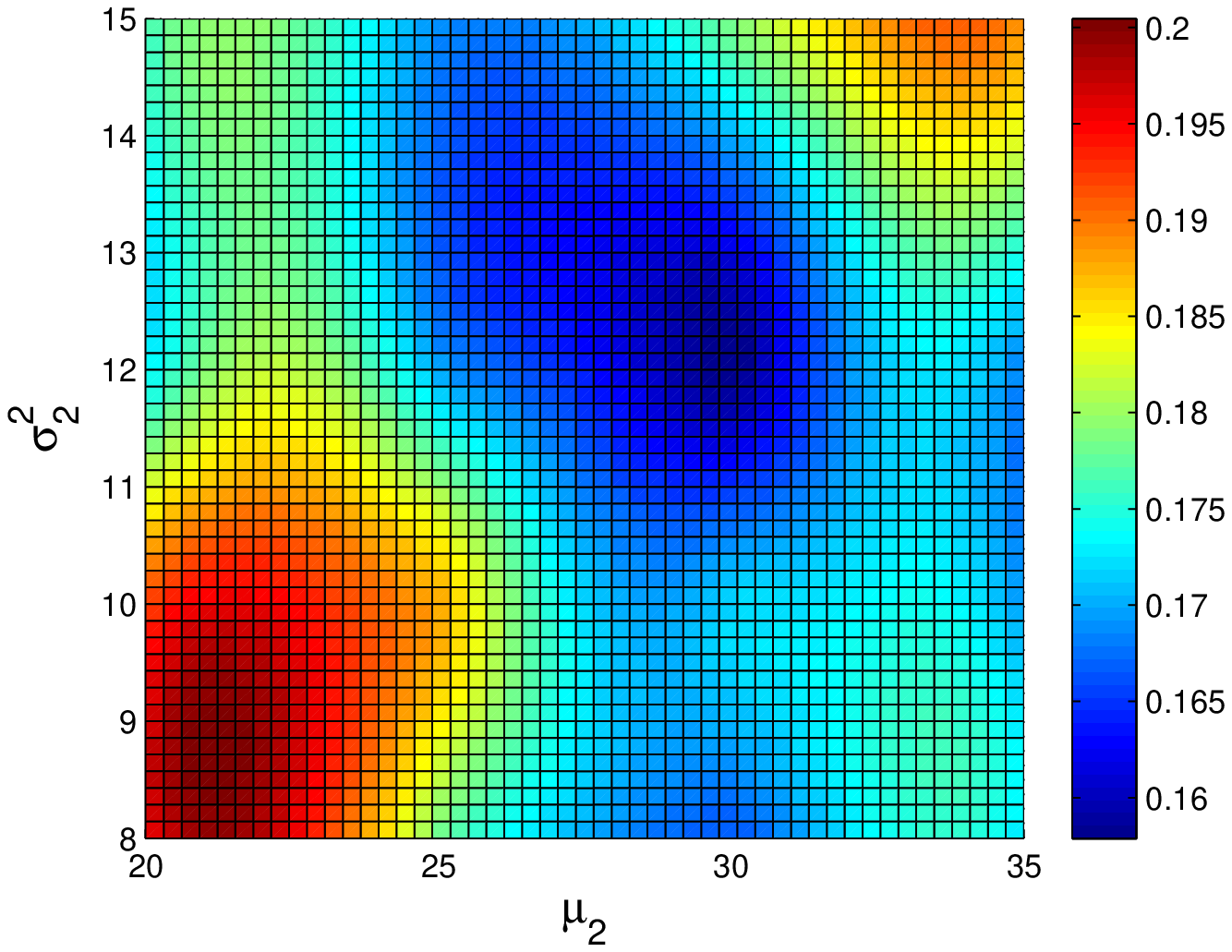}
    \label{fig-7b}
  }
\caption{Stability results of SAIoT-GR on Last.fm dataset.}
\label{fig-7 sensitivity last.fm}
\end{figure*}

\begin{figure*}[!ht]
\centering
  \subfigure[$\mu_1$ and $\sigma_1^2$]
  {
    \includegraphics[width=6cm]{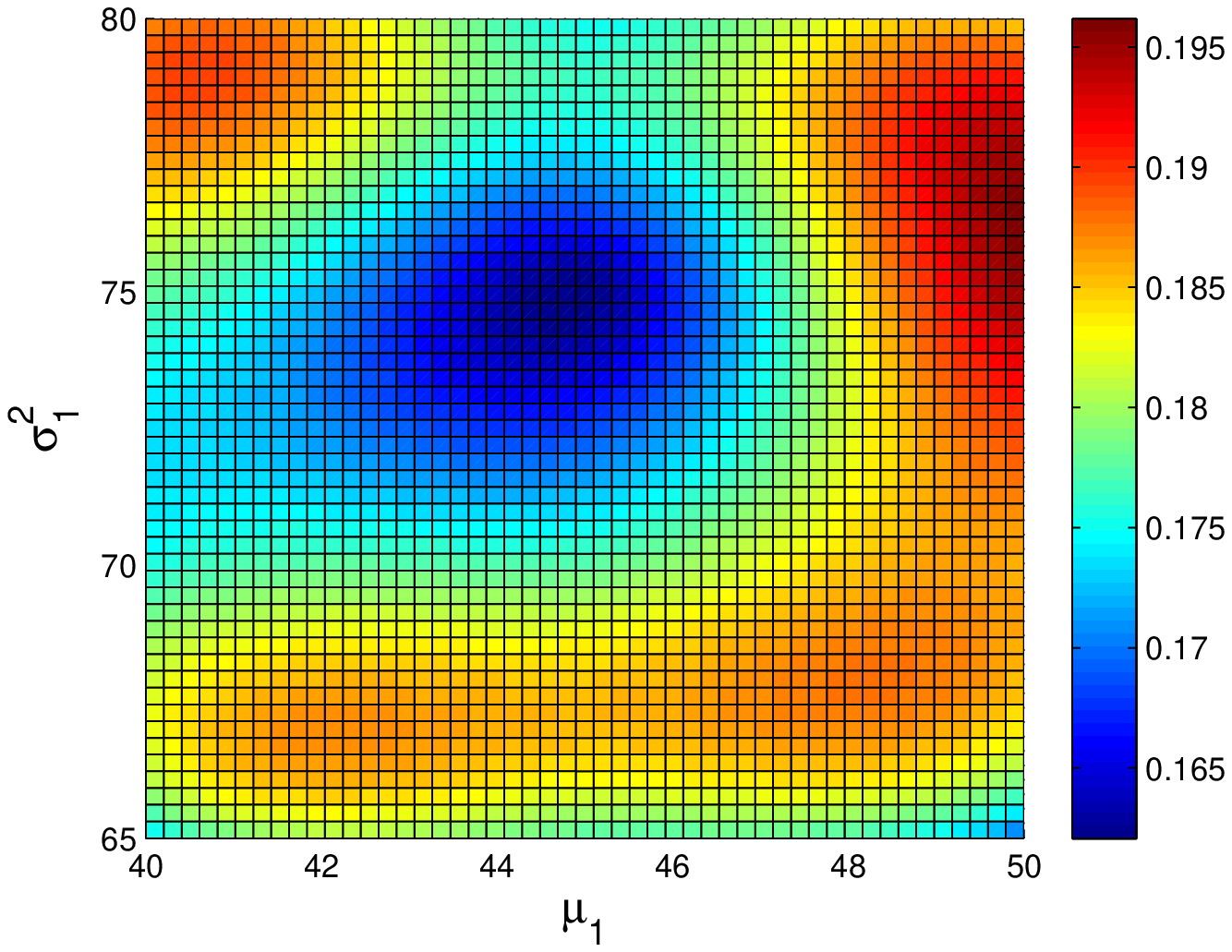}
    \label{fig-8a}
  }
  \subfigure[$\mu_2$ and $\sigma_2^2$]
  {
    \includegraphics[width=6cm]{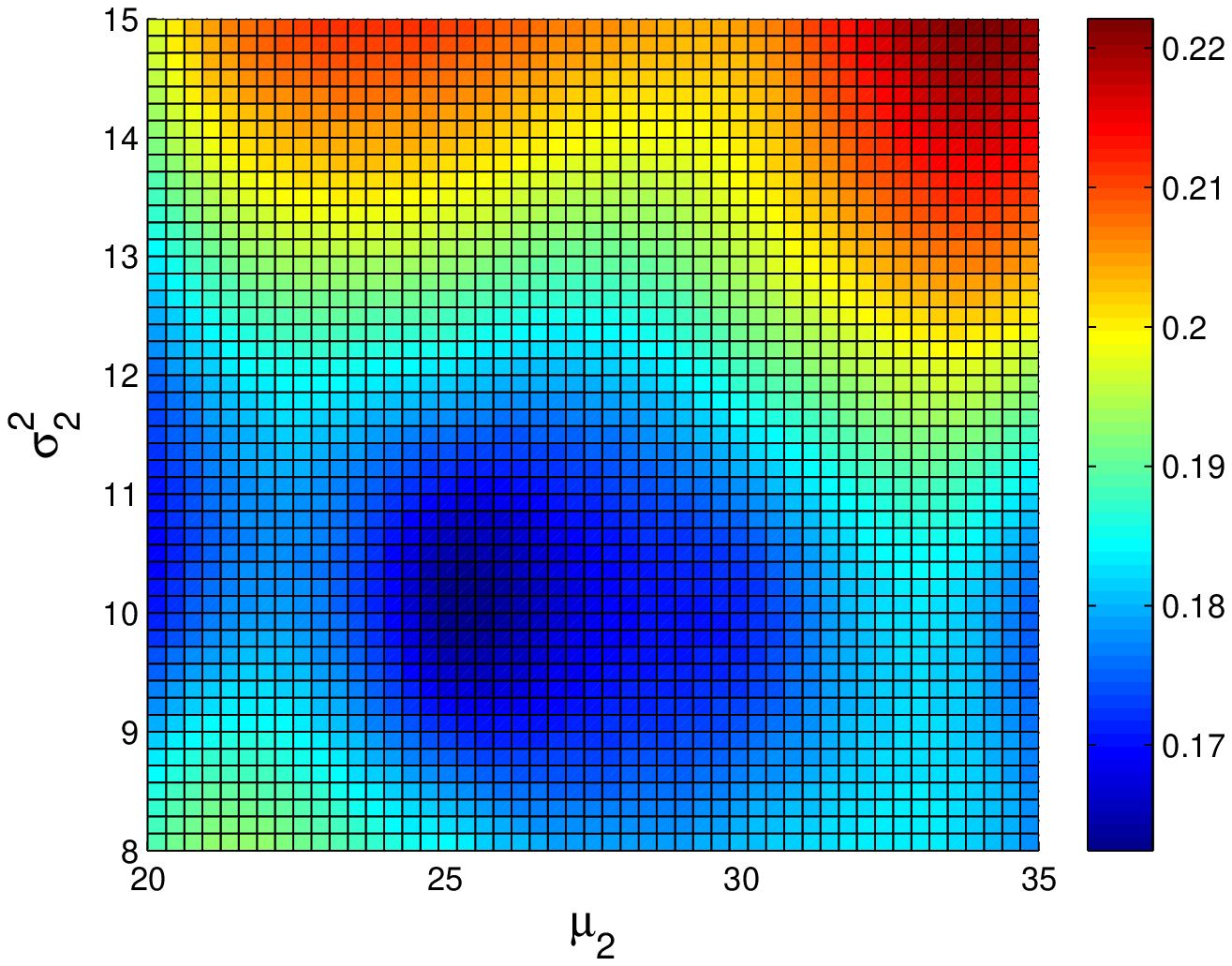}
    \label{fig-8b}
  }
\caption{Stability results of SAIoT-GR on Delicious dataset.}
\label{fig-8 sensitivity delicious}
\end{figure*}
\subsection{Results and Analysis}
Comparisons between predicted topic distributions and real
ones on the Last.fm dataset and the Delicious dataset,
are visualized on Fig.~\ref{fig-4 distribution comparision-last.fm}
and Fig.~\ref{fig-5 distribution comparision-delicious}, separately.
Naturally, the total number of topic indicators on two datasets
equals to 6 and 10. The Fig.~\ref{fig-4 distribution comparision-last.fm}
and Fig.~\ref{fig-5 distribution comparision-delicious} are
both made up of six subfigures, in which each one corresponds
to an experimental method. In six subfigures of
Fig.~\ref{fig-4 distribution comparision-last.fm}, the curves
in red color denote real values of topic distributions ranging
from topic 1 to topic 6, and the curves in blue color denote
predicted values of topic distributions ranging from topic 1
to topic 6. In six subfigures of Fig.~\ref{fig-5 distribution comparision-delicious}, the curves in green color denote real
values of topic distributions ranging from topic 1 to topic 6,
and the curves in pink color denote predicted values of topic
distributions ranging from topic 1 to topic 6.
It is revealed that distance measurement in terms of SAIoT-GR
is the smallest inside all the six methods. This reflects the
fact that performance of the SAIoT-GR is superior to ConfiMF
and Frequency. The ConfiMF is a popularity-based method that
investigates preference features from the perspective of
statistics, yet ignores complex group-level features inside
social networks. But these subfigures are not able to reveal
the comparisions between SAIoT-GR and the other several methods.
Thus the metrics are quantified to further compare performance
of SAIoT-GR and others.

TABLE~\ref{tab-2 six distances-last.fm} and
TABLE~\ref{tab-3 six distances-delicious} illustrate specific
values of six obtained evaluation metrics as experimental results.
They both have six columns that correspond to six distance
measurement, as well as six lines that correspond to six
experimental methods. Overall, the proposed
SAIoT-GR is about 4\% better than RandGroup, 6\% better than
FreGroup, 7\% better than ContextMF, 12\% better than ConfiMF and
15\% better than Frequency.
Although social information and contextual information are
also considered in ContextMF and RanGroup, the two methods
never deeply mining influential factors as the SAIoT-GR.
Differently, the SAIoT-GR tries to fuse information come from
multiple sources, and forms strong feature extraction for
the level of groups.
At the same time, it can be also observed that SAIoT-GR cannot acquire the best
experimental results all the time, such as results of Chebyshev distance on Last.fm
dataset. Two possible reasons can be speculated to explain this phenomenon. On the
one hand, some difference exists among computational methods of different distance
measurement criteria, so that it is difficult to guarantee that similar results can
be obtained by diverse criteria. On the other hand, SAIoT-GR is an unsupervised
learning method, which may bring about some uncertainty.

During previous experiments, prior parameters $\mu_1$, $\sigma_1^2$, $\mu_2$ and
$\sigma_2^2$ were set up empirically. As some unsupervised methods are susceptible
to initial value settings, an additional group of experiments
are conducted to testify stability for the proposed SAIoT-GR.
It should be noted that this group of experiments set no
comparisons between object method and baseline methods, just
performance of SAIoT-GR is singly evaluated. It is not required
to use all of the six distances in this group of experiments,
and just the most typical Euclidean distance is employed.
Fig.~\ref{fig-7 sensitivity last.fm} and
Fig.~\ref{fig-8 sensitivity delicious} illustrate the results
of parameter sensitivity testing on two datasets.
Inside these figures, most of the areas are in color of blue,
and no sharp change of colors is reflected.
It can be concluded from this group of experiments that the
SAIoT-GR is not only stable, but also robust.

\section{Conclusion}
Nowadays, recommender systems oriented to groups rather than individuals have received wide attention in academia. Existing technical approaches mostly utilized explicit feedbacks to develop GRSs. However, situations about implicit feedbacks which are common in the real-world are not well considered. Besides, the absence of online data management also acts as an obstacle to improve recommendation performance. To this end, this paper proposes SAIoT-GR, a CBN-based AIoT, to carry out group recommendations.
As for the hardware module, an exclusive IoT structure is developed as the bottom support platform. As for the software module, CBN model and non-cooperative game are can be introduced as algorithms. Such an AIoT architecture is able to maximize the advantages of the two modules. In addition, a large number of experiments are carried out to evaluate the performance of the SAIoT-GR in terms of efficiency and robustness. In the future, we are going to extend our SAIoT-GR to more complex scenarios, where the group’s preferences are not clear given the not enough explicit ratings or implicit feedback. Anoloty to cold-start users, such groups can be named as the cold-start groups. This is a also long-standing challenge in conventional RSs. We find a promising solution in \cite{r32, r33, r34}, which incorporates conversational interactions in RSs and demonstrates its efficiency and explainability over state-of-the-art methods. We hope that our SAIoT-GR will be able to make an explainable recommendations for cold-start groups by using conversational recommender strategies.





\ifCLASSOPTIONcaptionsoff
  \newpage
\fi

\end{document}